\documentclass[10pt,twocolumn,letterpaper]{article}

\usepackage{cvpr}
\usepackage{times}
\usepackage{epsfig}
\usepackage{graphicx}
\usepackage{amsmath}
\usepackage{amssymb}
\usepackage{subcaption}

\usepackage{url}
\usepackage{relsize}
\usepackage{xcolor,colortbl}
\usepackage{multirow}
\usepackage{wrapfig}
\usepackage{bbm}
\usepackage[labelfont=bf]{caption}
\usepackage{tabularx}

\usepackage[pagebackref=true,breaklinks=true,letterpaper=true,colorlinks,bookmarks=false]{hyperref}

\cvprfinalcopy % *** Uncomment this line for the final submission

\def\cvprPaperID{2411} % *** Enter the CVPR Paper ID here

\ifcvprfinal\fi
\begin{document}

%%%%%%%%% TITLE
\title{\vspace{-1cm}You Only Look Once: \\
Unified, Real-Time Object Detection\vspace{-.25cm}}

%\author{Joseph Redmon\\
%University of Washington\\
%{\tt\small pjreddie@cs.washington.edu}
% omit the following lines up until the closing `` }''.
%\and
%Santosh Divvala\\
%Allen Institute for Artificial Intelligence\\
%{\tt\small santoshd@allenai.org}
%\and
%Ross Girshick\\
%Facebook AI Research\\
%{\tt\small rbg@fb.com}
%\and
%Ali Farhadi\\
%University of Washington\\
%{\tt\small ali@cs.washington.edu}
%}

\author{Joseph Redmon$^*$, Santosh Divvala$^{* \dag}$, Ross Girshick$^\P$, Ali Farhadi$^{* \dag}$\\
\small{University of Washington$^*$, Allen Institute for AI$^\dag$, Facebook AI Research$^\P$}\\ \url{http://pjreddie.com/yolo/}}

\maketitle
%\thispagestyle{empty}

%%%%%%%%% ABSTRACT
\begin{abstract}
\vspace{-.25cm}
We present YOLO, a new approach to object detection. Prior work on object detection repurposes classifiers to perform detection. Instead, we frame object detection as a regression problem to spatially separated bounding boxes and associated class probabilities. A single neural network predicts bounding boxes and class probabilities directly from full images in one evaluation. Since the whole detection pipeline is a single network, it can be optimized end-to-end directly on detection performance.

Our unified architecture is extremely fast. Our base YOLO model processes images in real-time at 45 frames per second. A smaller version of the network, Fast YOLO, processes an astounding 155 frames per second while still achieving double the mAP of other real-time detectors. Compared to state-of-the-art detection systems, YOLO makes more localization errors but is less likely to predict false positives on background. Finally, YOLO learns very general representations of objects. It outperforms other detection methods, including DPM and R-CNN, when generalizing from natural images to other domains like artwork.
\end{abstract}

\section{Introduction}

Humans glance at an image and instantly know what objects are in the image, where they are, and how they interact. The human visual system is fast and accurate, allowing us to perform complex tasks like driving with little conscious thought. Fast, accurate algorithms for object detection would allow computers to drive cars without specialized sensors, enable assistive devices to convey real-time scene information to human users, and unlock the potential for general purpose, responsive robotic systems.
%
%Convolutional neural networks (CNNs) achieve impressive performance on classification tasks at real-time speeds \cite{DBLP:journals/corr/HeZR015}. Yet top object detection systems like R-CNN take seconds to process individual images and hallucinate objects in background noise. We believe these shortcomings result from how these systems approach object detection.

Current detection systems repurpose classifiers to perform detection. To detect an object, these systems take a classifier for that object and evaluate it at various locations and scales in a test image. Systems like deformable parts models (DPM) use a sliding window approach where the classifier is run at evenly spaced locations over the entire image \cite{lsvm-pami}.

More recent approaches like R-CNN use region proposal methods to first generate potential bounding boxes in an image and then run a classifier on these proposed boxes. After classification, post-processing is used to refine the bounding boxes, eliminate duplicate detections, and rescore the boxes based on other objects in the scene \cite{girshick2014rich}. These complex pipelines are slow and hard to optimize because each individual component must be trained separately.

\begin{figure}[t]
\begin{center}
        \includegraphics[width=\linewidth]{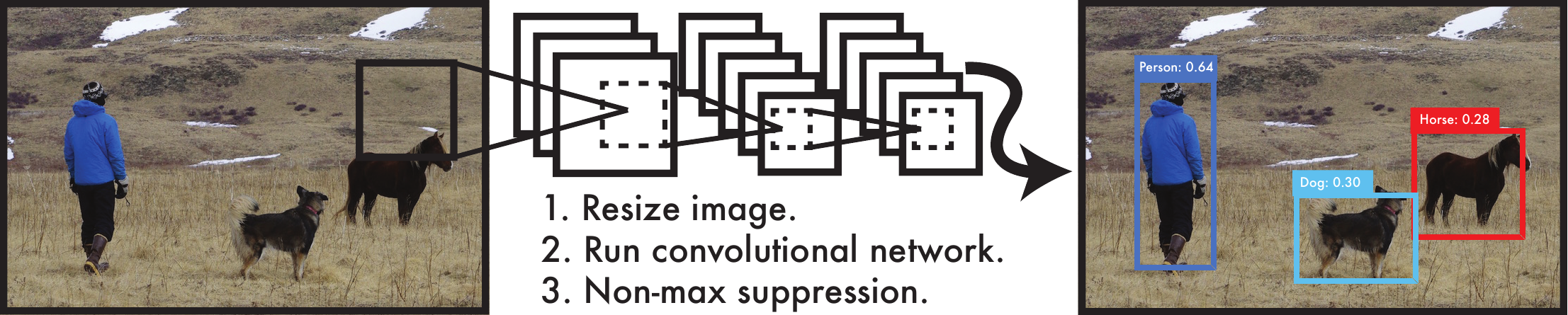}
\end{center}
   \caption{\small \textbf{The YOLO Detection System.} Processing images with YOLO is simple and straightforward. Our system (1) resizes the input image to $448 \times 448$, (2) runs a single convolutional network on the image, and (3) thresholds the resulting detections by the model's confidence.}
\label{system}
\end{figure}

%These region proposal techniques typically generate a few thousand potential boxes per image. Selective Search, the most common region proposal method, takes 1-2 seconds per image to generate these boxes \cite{uijlings2013selective}. The classifier then takes additional time to evaluate the proposals. The best performing systems require 2-40 seconds per image and even those optimized for speed do not achieve real-time performance. Additionally, even a highly accurate classifier will produce false positives when faced with so many proposals. When viewed out of context, small sections of background can resemble actual objects, causing detection errors.

%Finally, these detection pipelines rely on independent techniques at every stage that cannot be optimized jointly. A typical pipeline uses Selective Search for region proposals, a convolutional network for feature extraction, a collection of one-versus-all SVMs for classification, non-maximal suppression to reduce duplicates, and a linear model to adjust the final bounding box coordinates. Selective Search tries to maximize recall while the SVMs optimize for single class accuracy and the linear model learns from localization error.

We reframe object detection as a single regression problem, straight from image pixels to bounding box coordinates and class probabilities. Using our system, you only look once (YOLO) at an image to predict what objects are present and where they are.

YOLO is refreshingly simple: see Figure \ref{system}. A single convolutional network simultaneously predicts multiple bounding boxes and class probabilities for those boxes. YOLO trains on full images and directly optimizes detection performance. This unified model has several benefits over traditional methods of object detection.

First, YOLO is extremely fast. Since we frame detection as a regression problem we don't need a complex pipeline. We simply run our neural network on a new image at test time to predict detections. Our base network runs at 45 frames per second with no batch processing on a Titan X GPU and a fast version runs at more than 150 fps. This means we can process streaming video in real-time with less than 25 milliseconds of latency. Furthermore, YOLO achieves more than twice the mean average precision of other real-time systems. For a demo of our system running in real-time on a webcam please see our project webpage: \url{http://pjreddie.com/yolo/}.

Second, YOLO reasons globally about the image when making predictions. Unlike sliding window and region proposal-based techniques, YOLO sees the entire image during training and test time so it implicitly encodes contextual information about classes as well as their appearance. Fast R-CNN, a top detection method \cite{DBLP:journals/corr/Girshick15}, mistakes background patches in an image for objects because it can't see the larger context. YOLO makes less than half the number of background errors compared to Fast R-CNN.

Third, YOLO learns generalizable representations of objects. When trained on natural images and tested on artwork, YOLO outperforms top detection methods like DPM and R-CNN by a wide margin. Since YOLO is highly generalizable it is less likely to break down when applied to new domains or unexpected inputs.

YOLO still lags behind state-of-the-art detection systems in accuracy. While it can quickly identify objects in images it struggles to precisely localize some objects, especially small ones. We examine these tradeoffs further in our experiments.

All of our training and testing code is open source.
A variety of pretrained models are also available to download.

\section{Unified Detection}

We unify the separate components of object detection into a single neural network. Our network uses features from the entire image to predict each bounding box. It also predicts all bounding boxes across all classes for an image simultaneously. This means our network reasons globally about the full image and all the objects in the image. The YOLO design enables end-to-end training and real-time speeds while maintaining high average precision.

Our system divides the input image into an $S \times S$ grid. If the center of an object falls into a grid cell, that grid cell is responsible for detecting that object.

Each grid cell predicts $B$ bounding boxes and confidence scores for those boxes. These confidence scores reflect how confident the model is that the box contains an object and also how accurate it thinks the box is that it predicts. Formally we define confidence as $\Pr(\textrm{Object}) * \textrm{IOU}_{\textrm{pred}}^{\textrm{truth}}$. If no object exists in that cell, the confidence scores should be zero. Otherwise we want the confidence score to equal the intersection over union (IOU) between the predicted box and the ground truth.

Each bounding box consists of 5 predictions: $x$, $y$, $w$, $h$, and confidence. The $(x,y)$ coordinates represent the center of the box relative to the bounds of the grid cell. The width and height are predicted relative to the whole image. Finally the confidence prediction represents the IOU between the predicted box and any ground truth box.

Each grid cell also predicts $C$ conditional class probabilities, $\Pr(\textrm{Class}_i | \textrm{Object})$. These probabilities are conditioned on the grid cell containing an object. We only predict one set of class probabilities per grid cell, regardless of the number of boxes $B$.

At test time we multiply the conditional class probabilities and the individual box confidence predictions,
\begin{equation}
\scriptsize
\Pr(\textrm{Class}_i | \textrm{Object}) * \Pr(\textrm{Object}) * \textrm{IOU}_{\textrm{pred}}^{\textrm{truth}} = \Pr(\textrm{Class}_i)*\textrm{IOU}_{\textrm{pred}}^{\textrm{truth}}
\end{equation}
which gives us class-specific confidence scores for each box. These scores encode both the probability of that class appearing in the box and how well the predicted box fits the object.

\begin{figure}[h]
\begin{center}
    \includegraphics[width=\linewidth]{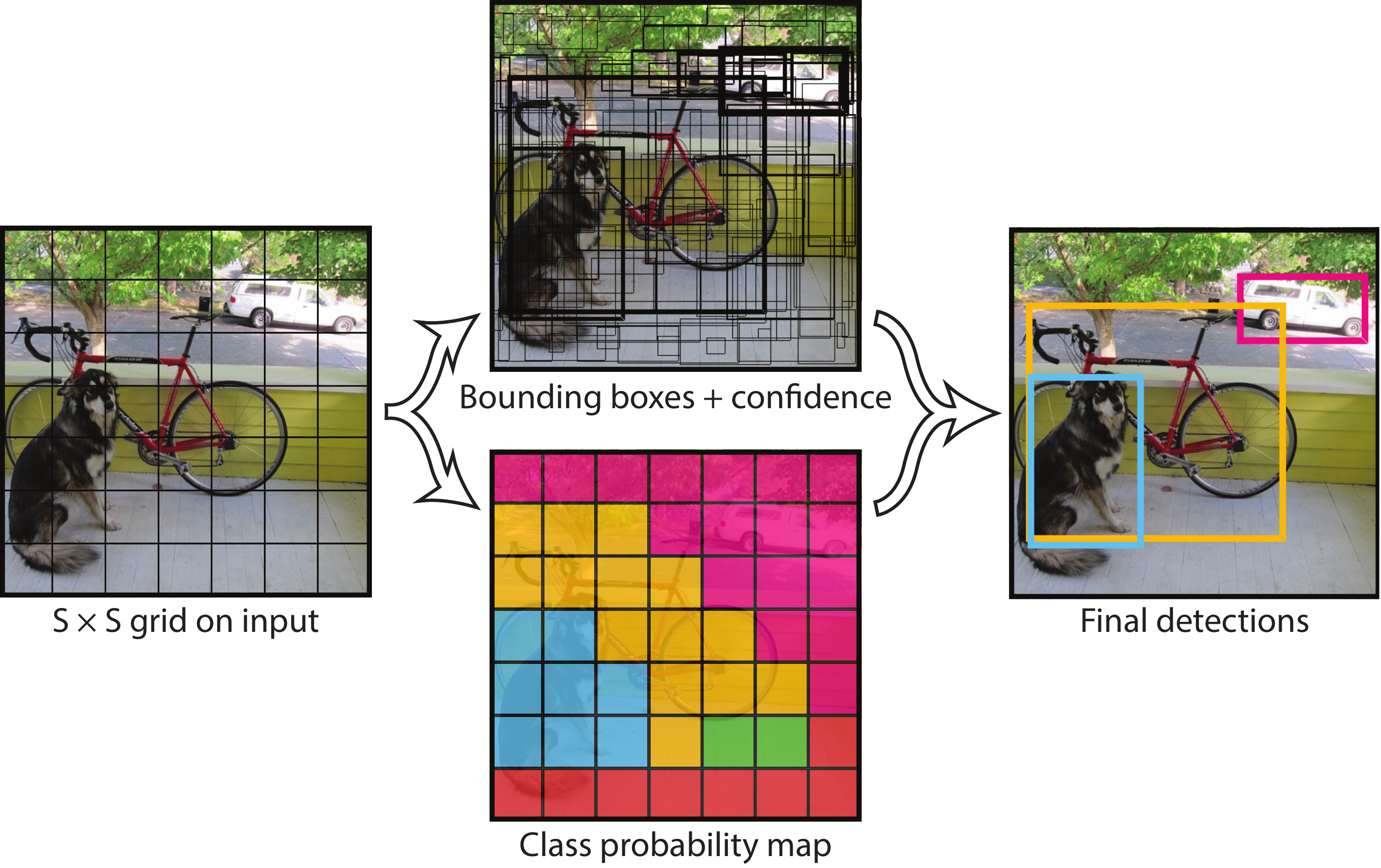}
\end{center}
   \caption{\small \textbf{The Model.} Our system models detection as a regression problem. It divides the image into an $S \times S$ grid and for each grid cell predicts $B$ bounding boxes, confidence for those boxes, and $C$ class probabilities. These predictions are encoded as an $S \times S \times (B*5 + C)$ tensor.}
\label{model}
\end{figure}

For evaluating YOLO on \textsc{Pascal} VOC, we use $S=7$, $B=2$. \textsc{Pascal} VOC has 20 labelled classes so $C=20$. Our final prediction is a $7 \times 7 \times 30$ tensor.

\subsection{Network Design}

   \begin{figure*}[t]
      \centering
        \includegraphics[width=.8\linewidth]{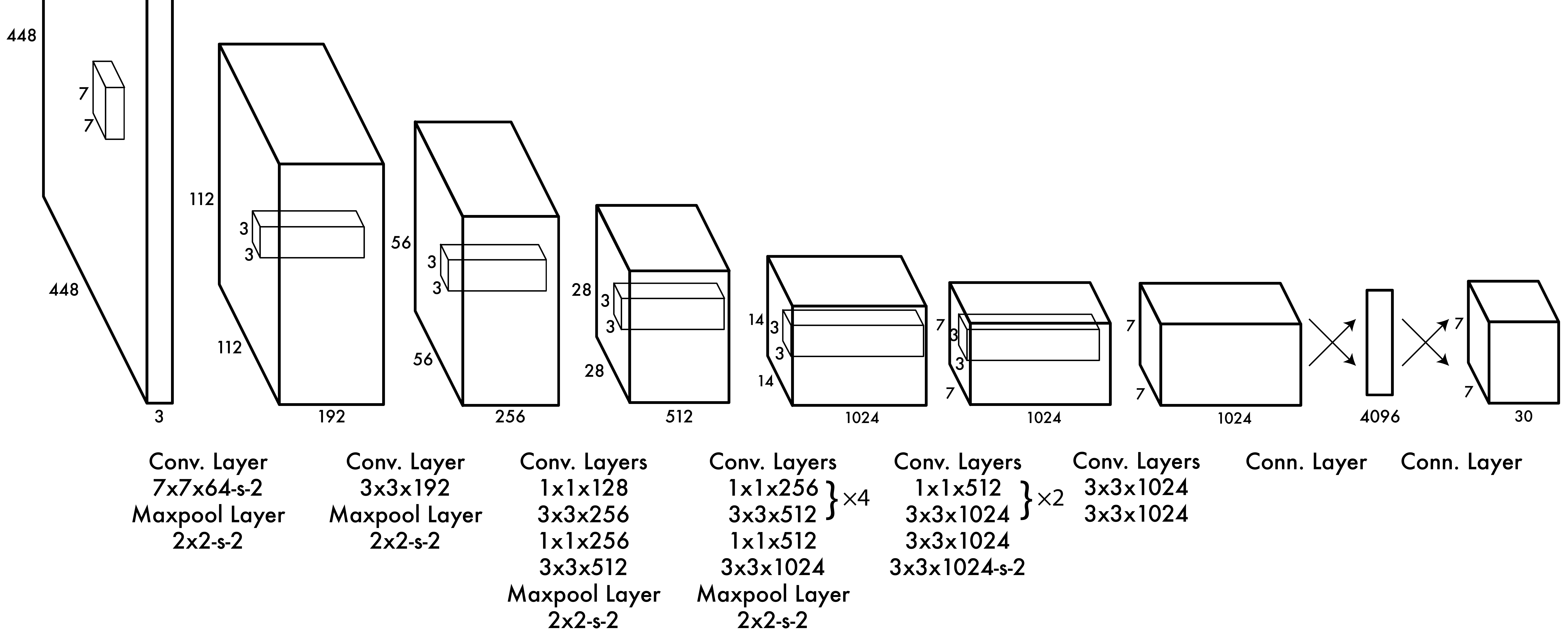}
      \caption{\small \textbf{The Architecture.} Our detection network has 24 convolutional layers followed by 2 fully connected layers. Alternating $1 \times 1$ convolutional layers reduce the features space from preceding layers. We pretrain the convolutional layers on the ImageNet classification task at half the resolution ($224 \times 224$ input image) and then double the resolution for detection.}
      \label{net}
   \end{figure*}

We implement this model as a convolutional neural network and evaluate it on the \textsc{Pascal} VOC detection dataset \cite{Everingham15}. The initial convolutional layers of the network extract features from the image while the fully connected layers predict the output probabilities and coordinates.

Our network architecture is inspired by the GoogLeNet model for image classification \cite{DBLP:journals/corr/SzegedyLJSRAEVR14}. Our network has 24 convolutional layers followed by 2 fully connected layers. Instead of the inception modules used by GoogLeNet, we simply use $1 \times 1$ reduction layers followed by $3 \times 3$ convolutional layers, similar to Lin et al \cite{DBLP:journals/corr/LinCY13}. The full network is shown in Figure \ref{net}.

We also train a fast version of YOLO designed to push the boundaries of fast object detection. Fast YOLO uses a neural network with fewer convolutional layers (9 instead of 24) and fewer filters in those layers. Other than the size of the network, all training and testing parameters are the same between YOLO and Fast YOLO.

The final output of our network is the $7 \times 7 \times 30$ tensor of predictions.

\subsection{Training}
%
%   \begin{figure}[t]
%      \centering
%        \includegraphics[width=.8\linewidth]{transform}
%      \caption{\textbf{The Coordinate System.} This image shows an example transformation from bounding box coordinates to the coordinates used by our output layer. Left: We normalize width and height by the width and height of the image. Middle: We find the appropriate grid cell based on the center of the bounding box. Right: We calculate the center coordinate offsets relative to the position of the grid cell.}
%      \label{transform}
%   \end{figure}

We pretrain our convolutional layers on the ImageNet 1000-class competition dataset \cite{ILSVRC15}. For pretraining we use the first 20 convolutional layers from Figure \ref{net} followed by a average-pooling layer and a fully connected layer. We train this network for approximately a week and achieve a single crop top-5 accuracy of 88\% on the ImageNet 2012 validation set, comparable to the GoogLeNet models in Caffe's Model Zoo \cite{zoo}. We use the Darknet framework for all training and inference \cite{darknet13}.

We then convert the model to perform detection. Ren et al. show that adding both convolutional and connected layers to pretrained networks can improve performance \cite{DBLP:journals/corr/RenHGZ015}. Following their example, we add four convolutional layers and two fully connected layers with randomly initialized weights. Detection often requires fine-grained visual information so we increase the input resolution of the network from $224 \times 224$ to $448 \times 448$.

Our final layer predicts both class probabilities and bounding box coordinates. We normalize the bounding box width and height by the image width and height so that they fall between 0 and 1. We parametrize the bounding box $x$ and $y$ coordinates to be offsets of a particular grid cell location so they are also bounded between 0 and 1.

We use a linear activation function for the final layer and all other layers use the following leaky rectified linear activation:

\begin{equation}
\phi(x) =
\begin{cases}
    x, & \text{if } x > 0\\
    0.1x, & \text{otherwise}
    \end{cases}
\end{equation}

We optimize for sum-squared error in the output of our model. We use sum-squared error because it is easy to optimize, however it does not perfectly align with our goal of maximizing average precision. It weights localization error equally with classification error which may not be ideal. Also, in every image many grid cells do not contain any object. This pushes the ``confidence'' scores of those cells towards zero, often overpowering the gradient from cells that do contain objects. This can lead to model instability, causing training to diverge early on.

To remedy this, we increase the loss from bounding box coordinate predictions and decrease the loss from confidence predictions for boxes that don't contain objects. We use two parameters, $\lambda_\textrm{coord}$ and $\lambda_\textrm{noobj}$ to accomplish this. We set $\lambda_\textrm{coord} = 5$ and $\lambda_\textrm{noobj} = .5$.

Sum-squared error also equally weights errors in large boxes and small boxes. Our error metric should reflect that small deviations in large boxes matter less than in small boxes. To partially address this we predict the square root of the bounding box width and height instead of the width and height directly. 
%Figure \ref{transform} shows an example transformation of a bounding box from image coordinates to the coordinates used by YOLO.

YOLO predicts multiple bounding boxes per grid cell. At training time we only want one bounding box predictor to be responsible for each object. We assign one predictor to be ``responsible'' for predicting an object based on which prediction has the highest current IOU with the ground truth. This leads to specialization between the bounding box predictors. Each predictor gets better at predicting certain sizes, aspect ratios, or classes of object, improving overall recall.

During training we optimize the following, multi-part loss function:
\scriptsize
\begin{multline}
\lambda_\textbf{coord}
\sum_{i = 0}^{S^2}
    \sum_{j = 0}^{B}
     \mathlarger{\mathbbm{1}}_{ij}^{\text{obj}}
            \left[
            \left(
                x_i - \hat{x}_i
            \right)^2 +
            \left(
                y_i - \hat{y}_i
            \right)^2
            \right]
\\
+ \lambda_\textbf{coord} 
\sum_{i = 0}^{S^2}
    \sum_{j = 0}^{B}
         \mathlarger{\mathbbm{1}}_{ij}^{\text{obj}}
         \left[
        \left(
            \sqrt{w_i} - \sqrt{\hat{w}_i}
        \right)^2 +
        \left(
            \sqrt{h_i} - \sqrt{\hat{h}_i}
        \right)^2
        \right]
\\
+ \sum_{i = 0}^{S^2}
    \sum_{j = 0}^{B}
        \mathlarger{\mathbbm{1}}_{ij}^{\text{obj}}
        \left(
            C_i - \hat{C}_i
        \right)^2
\\
+ \lambda_\textrm{noobj}
\sum_{i = 0}^{S^2}
    \sum_{j = 0}^{B}
    \mathlarger{\mathbbm{1}}_{ij}^{\text{noobj}}
        \left(
            C_i - \hat{C}_i
        \right)^2
\\
+ \sum_{i = 0}^{S^2}
\mathlarger{\mathbbm{1}}_i^{\text{obj}}
    \sum_{c \in \textrm{classes}}
        \left(
            p_i(c) - \hat{p}_i(c)
        \right)^2
\end{multline}
\normalsize
where $\mathbbm{1}_i^{\text{obj}}$ denotes if object appears in cell $i$ and $\mathbbm{1}_{ij}^{\text{obj}}$ denotes that the $j$th bounding box predictor in cell $i$ is ``responsible'' for that prediction.

Note that the loss function only penalizes classification error if an object is present in that grid cell (hence the conditional class probability discussed earlier). It also only penalizes bounding box coordinate error if that predictor is ``responsible'' for the ground truth box (i.e. has the highest IOU of any predictor in that grid cell).

We train the network for about 135 epochs on the training and validation data sets from \textsc{Pascal} VOC 2007 and 2012. When testing on 2012 we also include the VOC 2007 test data for training. Throughout training we use a batch size of 64, a momentum of $0.9$ and a decay of $0.0005$.

Our learning rate schedule is as follows: For the first epochs we slowly raise the learning rate from $10^{-3}$ to $10^{-2}$. If we start at a high learning rate our model often diverges due to unstable gradients. We continue training with $10^{-2}$ for 75 epochs, then $10^{-3}$ for 30 epochs, and finally $10^{-4}$ for 30 epochs.

To avoid overfitting we use dropout and extensive data augmentation. A dropout layer with rate~=~.5 after the first connected layer prevents co-adaptation between layers \cite{hinton2012improving}. For data augmentation we introduce random scaling and translations of up to 20\% of the original image size. We also randomly adjust the exposure and saturation of the image by up to a factor of $1.5$ in the HSV color space.

\subsection{Inference}

Just like in training, predicting detections for a test image only requires one network evaluation. On \textsc{Pascal} VOC the network predicts 98 bounding boxes per image and class probabilities for each box. YOLO is extremely fast at test time since it only requires a single network evaluation, unlike classifier-based methods.

The grid design enforces spatial diversity in the bounding box predictions. Often it is clear which grid cell an object falls in to and the network only predicts one box for each object. However, some large objects or objects near the border of multiple cells can be well localized by multiple cells. Non-maximal suppression can be used to fix these multiple detections. While not critical to performance as it is for R-CNN or DPM, non-maximal suppression adds 2-3\% in mAP.

\subsection{Limitations of YOLO}

YOLO imposes strong spatial constraints on bounding box predictions since each grid cell only predicts two boxes and can only have one class. This spatial constraint limits the number of nearby objects that our model can predict. Our model struggles with small objects that appear in groups, such as flocks of birds.

Since our model learns to predict bounding boxes from data, it struggles to generalize to objects in new or unusual aspect ratios or configurations. Our model also uses relatively coarse features for predicting bounding boxes since our architecture has multiple downsampling layers from the input image.

Finally, while we train on a loss function that approximates detection performance, our loss function treats errors the same in small bounding boxes versus large bounding boxes. A small error in a large box is generally benign but a small error in a small box has a much greater effect on IOU. Our main source of error is incorrect localizations.

\section{Comparison to Other Detection Systems}

Object detection is a core problem in computer vision. Detection pipelines generally start by extracting a set of robust features from input images (Haar \cite{papageorgiou1998general}, SIFT \cite{lowe1999object}, HOG \cite{dalal2005histograms}, convolutional features \cite{donahue2013decaf}). Then, classifiers \cite{viola2001robust,lienhart2002extended,girshick2014rich,lsvm-pami} or localizers \cite{blaschko2008learning,DBLP:journals/corr/SermanetEZMFL13} are used to identify objects in the feature space. These classifiers or localizers are run either in sliding window fashion over the whole image or on some subset of regions in the image \cite{uijlings2013selective,gould2009region,zitnick2014edge}. We compare the YOLO detection system to several top detection frameworks, highlighting key similarities and differences.

\textbf{Deformable parts models.} Deformable parts models (DPM) use a sliding window approach to object detection \cite{lsvm-pami}. DPM uses a disjoint pipeline to extract static features, classify regions, predict bounding boxes for high scoring regions, etc. Our system replaces all of these disparate parts with a single convolutional neural network. The network performs feature extraction, bounding box prediction, non-maximal suppression, and contextual reasoning all concurrently. Instead of static features, the network trains the features in-line and optimizes them for the detection task. Our unified architecture leads to a faster, more accurate model than DPM.

\textbf{R-CNN.} R-CNN and its variants use region proposals instead of sliding windows to find objects in images. Selective Search \cite{uijlings2013selective} generates potential bounding boxes, a convolutional network extracts features, an SVM scores the boxes, a linear model adjusts the bounding boxes, and non-max suppression eliminates duplicate detections. Each stage of this complex pipeline must be precisely tuned independently and the resulting system is very slow, taking more than 40 seconds per image at test time \cite{DBLP:journals/corr/Girshick15}.

YOLO shares some similarities with R-CNN. Each grid cell proposes potential bounding boxes and scores those boxes using convolutional features. However, our system puts spatial constraints on the grid cell proposals which helps mitigate multiple detections of the same object. Our system also proposes far fewer bounding boxes, only 98 per image compared to about 2000 from Selective Search. Finally, our system combines these individual components into a single, jointly optimized model.

\textbf{Other Fast Detectors} Fast and Faster R-CNN focus on speeding up the R-CNN framework by sharing computation and using neural networks to propose regions instead of Selective Search \cite{DBLP:journals/corr/Girshick15} \cite{ren2015faster}. While they offer speed and accuracy improvements over R-CNN, both still fall short of real-time performance.

Many research efforts focus on speeding up the DPM pipeline \cite{sadeghi201430hz} \cite{yan2014fastest} \cite{dean2013fast}. They speed up HOG computation, use cascades, and push computation to GPUs. However, only 30Hz DPM \cite{sadeghi201430hz} actually runs in real-time.

Instead of trying to optimize individual components of a large detection pipeline, YOLO throws out the pipeline entirely and is fast by design.

Detectors for single classes like faces or people can be highly optimized since they have to deal with much less variation \cite{viola2004robust}. YOLO is a general purpose detector that learns to detect a variety of objects simultaneously.

\textbf{Deep MultiBox.} Unlike R-CNN, Szegedy et al. train a convolutional neural network to predict regions of interest \cite{erhan2014scalable} instead of using Selective Search. MultiBox can also perform single object detection by replacing the confidence prediction with a single class prediction. However, MultiBox cannot perform general object detection and is still just a piece in a larger detection pipeline, requiring further image patch classification. Both YOLO and MultiBox use a convolutional network to predict bounding boxes in an image but YOLO is a complete detection system.

\textbf{OverFeat.} Sermanet et al. train a convolutional neural network to perform localization and adapt that localizer to perform detection \cite{DBLP:journals/corr/SermanetEZMFL13}. OverFeat efficiently performs sliding window detection but it is still a disjoint system. OverFeat optimizes for localization, not detection performance. Like DPM, the localizer only sees local information when making a prediction. OverFeat cannot reason about global context and thus requires significant post-processing to produce coherent detections.

\textbf{MultiGrasp.} Our work is similar in design to work on grasp detection by Redmon et al \cite{DBLP:journals/corr/RedmonA14}. Our grid approach to bounding box prediction is based on the MultiGrasp system for regression to grasps. However, grasp detection is a much simpler task than object detection. MultiGrasp only needs to predict a single graspable region for an image containing one object. It doesn't have to estimate the size, location, or boundaries of the object or predict it's class, only find a region suitable for grasping. YOLO predicts both bounding boxes and class probabilities for multiple objects of multiple classes in an image.

\section{Experiments}

First we compare YOLO with other real-time detection systems on \textsc{Pascal} VOC 2007. To understand the differences between YOLO and R-CNN variants we explore the errors on VOC 2007 made by YOLO and Fast R-CNN, one of the highest performing versions of R-CNN \cite{DBLP:journals/corr/Girshick15}. Based on the different error profiles we show that YOLO can be used to rescore Fast R-CNN detections and reduce the errors from background false positives, giving a significant performance boost. We also present VOC 2012 results and compare mAP to current state-of-the-art methods. Finally, we show that YOLO generalizes to new domains better than other detectors on two artwork datasets.

\subsection{Comparison to Other Real-Time Systems}

Many research efforts in object detection focus on making standard detection pipelines fast. \cite{dean2013fast} \cite{yan2014fastest} \cite{sadeghi201430hz} \cite{DBLP:journals/corr/Girshick15} \cite{he2014spatial} \cite{ren2015faster} However, only Sadeghi et al. actually produce a detection system that runs in real-time (30 frames per second or better) \cite{sadeghi201430hz}. We compare YOLO to their GPU implementation of DPM which runs either at 30Hz or 100Hz. While the other efforts don't reach the real-time milestone we also compare their relative mAP and speed to examine the accuracy-performance tradeoffs available in object detection systems.

\begin{table}[h]
\begin{center}
\begin{tabular}{lrrr}
Real-Time Detectors & Train & mAP & FPS\\
\hline
100Hz DPM \cite{sadeghi201430hz}& 2007 & 16.0 & 100\\
30Hz DPM \cite{sadeghi201430hz} & 2007 & 26.1 & 30 \\
Fast YOLO & 2007+2012 & 52.7 & \textbf{155} \\
YOLO & 2007+2012 & \textbf{63.4} & 45 \\
\hline
\hline
Less Than Real-Time &  & \\
\hline
Fastest DPM \cite{yan2014fastest} & 2007 & 30.4 & 15 \\
R-CNN Minus R \cite{lenc2015r} & 2007 & 53.5 & 6 \\
Fast R-CNN \cite{DBLP:journals/corr/Girshick15}& 2007+2012 & 70.0 & 0.5 \\
Faster R-CNN VGG-16\cite{ren2015faster}& 2007+2012 & 73.2 & 7 \\
Faster R-CNN ZF \cite{ren2015faster}& 2007+2012 & 62.1 & 18 \\
YOLO VGG-16 & 2007+2012 & 66.4 & 21 \\
\end{tabular}
\end{center}
\caption{\small \textbf{Real-Time Systems on \textsc{Pascal} VOC 2007.} Comparing the performance and speed of fast detectors. Fast YOLO is the fastest detector on record for \textsc{Pascal} VOC detection and is still twice as accurate as any other real-time detector. YOLO is 10 mAP more accurate than the fast version while still well above real-time in speed.}
\label{timing}
\end{table}

Fast YOLO is the fastest object detection method on \textsc{Pascal}; as far as we know, it is the fastest extant object detector. With $52.7\%$ mAP, it is more than twice as accurate as prior work on real-time detection. YOLO pushes mAP to $63.4\%$ while still maintaining real-time performance.

We also train YOLO using VGG-16. This model is more accurate but also significantly slower than YOLO. It is useful for comparison to other detection systems that rely on VGG-16 but since it is slower than real-time the rest of the paper focuses on our faster models.

Fastest DPM effectively speeds up DPM without sacrificing much mAP but it still misses real-time performance by a factor of 2 \cite{yan2014fastest}. It also is limited by DPM's relatively low accuracy on detection compared to neural network approaches.

R-CNN minus R replaces Selective Search with static bounding box proposals \cite{lenc2015r}. While it is much faster than R-CNN, it still falls short of real-time and takes a significant accuracy hit from not having good proposals. 

Fast R-CNN speeds up the classification stage of R-CNN but it still relies on selective search which can take around 2 seconds per image to generate bounding box proposals. Thus it has high mAP but at $0.5$ fps it is still far from real-time.

The recent Faster R-CNN replaces selective search with a neural network to propose bounding boxes, similar to Szegedy et al. \cite{erhan2014scalable} In our tests, their most accurate model achieves 7 fps while a smaller, less accurate one runs at 18 fps. The VGG-16 version of Faster R-CNN is 10 mAP higher but is also 6 times slower than YOLO. The Zeiler-Fergus Faster R-CNN is only 2.5 times slower than YOLO but is also less accurate.

\subsection{VOC 2007 Error Analysis}
\label{error}

To further examine the differences between YOLO and state-of-the-art detectors, we look at a detailed breakdown of results on VOC 2007. We compare YOLO to Fast R-CNN since Fast R-CNN is one of the highest performing detectors on \textsc{Pascal} and it's detections are publicly available.

We use the methodology and tools of Hoiem et al. \cite{hoiem2012diagnosing} For each category at test time we look at the top N predictions for that category. Each prediction is either correct or it is classified based on the type of error:

\begin{itemize}
\itemsep0em
\item Correct: correct class and $\textrm{IOU} > .5$
\item Localization: correct class, $.1 < \textrm{IOU} < .5$
\item Similar: class is similar, $\textrm{IOU} > .1$
\item Other: class is wrong, $\textrm{IOU} > .1$
\item Background: $\textrm{IOU} < .1$ for any object
\end{itemize}

\begin{figure}[t]
      \centering
        \includegraphics[width=\linewidth]{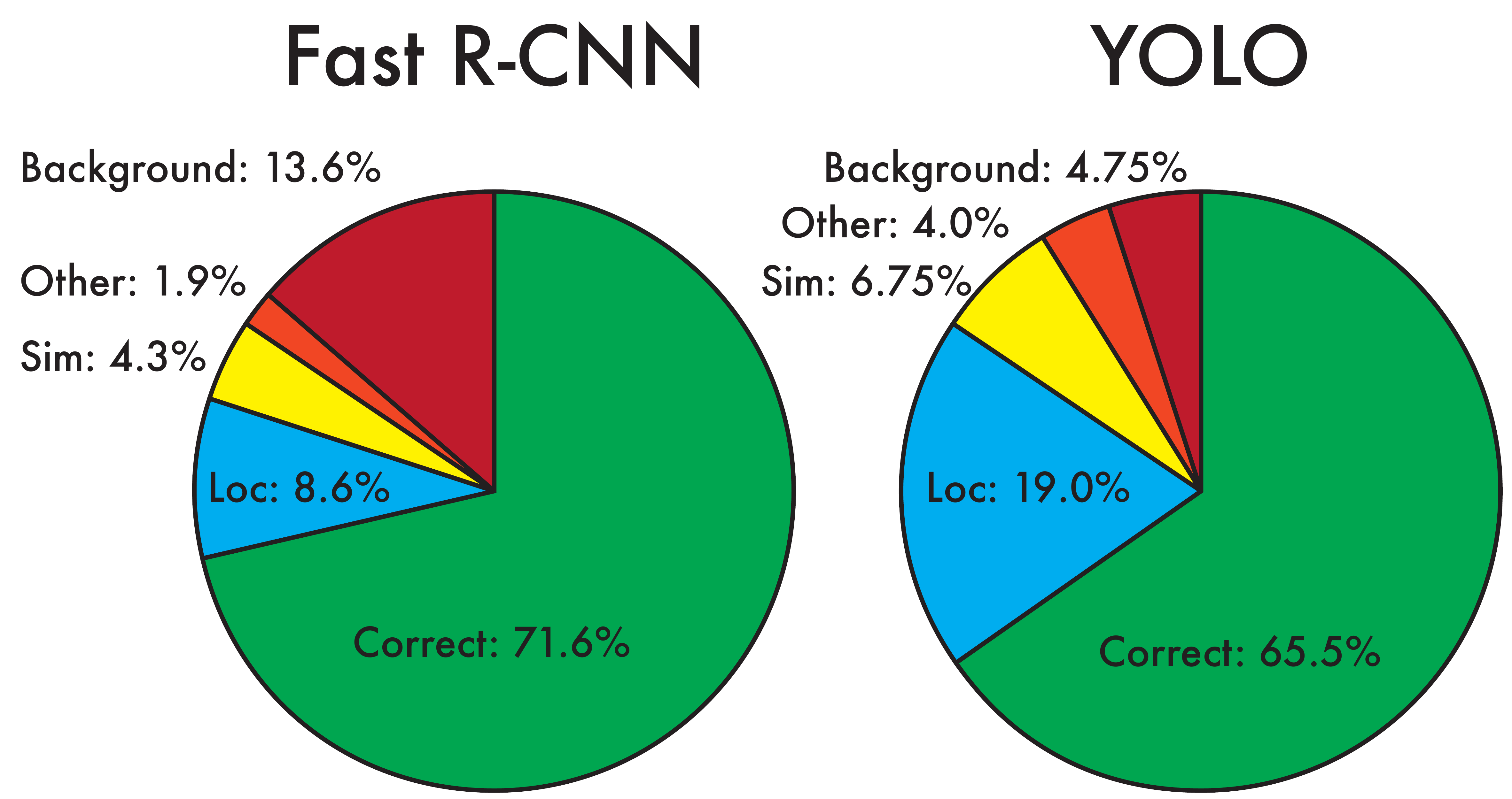}
      \caption{\small \textbf{Error Analysis: Fast R-CNN vs. YOLO} These charts show the percentage of localization and background errors in the top N detections for various categories (N = \# objects in that category).}
      \label{errors}
   \end{figure}

Figure \ref{errors} shows the breakdown of each error type averaged across all 20 classes.

%An object detector must have high recall for objects in the test set to obtain high performance. Our model imposes spatial constraints on bounding box predictions which limits recall on small objects that are close together. We examine how detrimental this is in practice by calculating our highest possible recall assuming perfect coordinate prediction. Under this assumption, our model can achieve a 93.1\% recall for objects in the VOC 2007 test set. This is lower than Selective Search (98.0\% \cite{uijlings2013selective}) but still relatively high.

%   \begin{figure*}[t]
%      \centering
%        \includegraphics[width=\linewidth]{sheep}
%      \caption{\small \textbf{Top 10 False Positives for Sheep: Fast R-CNN and YOLO.} These images show the top false detections on VOC 2007 for the sheep category. Fast R-CNN commonly mistakes patches of grass or rocks for sheep. YOLO tends to mistake other livestock as sheep as well as other objects that are simply nearby sheep (dogs, boulders, or people).}
%      \label{sheep}
%   \end{figure*}

YOLO struggles to localize objects correctly. Localization errors account for more of YOLO's errors than all other sources combined. Fast R-CNN makes much fewer localization errors but far more background errors. 13.6\% of it's top detections are false positives that don't contain any objects. Fast R-CNN is almost 3x more likely to predict background detections than YOLO.

%Figure \ref{sheep} shows the top 10 false positives for sheep on VOC 2007 for Fast R-CNN and YOLO. Half of Fast R-CNN's top errors are background detections; it detects patches of grass, small rocks, or hay bales as sheep.

%By contrast, YOLO only makes one background detection, on a large boulder next to some sheep. Apart from localization errors, YOLO mistakes other livestock as sheep, as well as other objects that just happen to be near sheep, like dogs or people.

\subsection{Combining Fast R-CNN and YOLO}

\begin{table}[b]
\begin{center}
\begin{tabular}{lrrr}
 & mAP & Combined & Gain \\
\hline
Fast R-CNN & 71.8 & - & - \\
\hline
Fast R-CNN (2007 data) & \textbf{66.9} & 72.4 & .6  \\
Fast R-CNN (VGG-M) & 59.2 & 72.4 & .6 \\
Fast R-CNN (CaffeNet) & 57.1  & 72.1 & .3\\
YOLO & 63.4  & \textbf{75.0} & \textbf{3.2}\\
\end{tabular}
\end{center}
\caption{\small \textbf{Model combination experiments on VOC 2007.} We examine the effect of combining various models with the best version of Fast R-CNN. Other versions of Fast R-CNN provide only a small benefit while YOLO provides a significant performance boost.}
\label{combine}
\end{table}

\begin{table*}[t]
\scriptsize
\definecolor{Gray}{gray}{0.85}
\newcolumntype{Y}{>{\centering\arraybackslash}X}
\begin{center}
\tabcolsep=0.11cm
\begin{tabularx}{\linewidth}{@{}l|Y|Y Y Y Y Y Y Y Y Y Y Y Y Y Y Y Y Y Y Y Y}
\textbf{VOC 2012 test} & mAP & aero & bike & bird & boat & bottle & bus & car & cat & chair & cow & table & dog & horse & mbike & person & plant & sheep & sofa & train & tv \\
\hline
MR\_CNN\_MORE\_DATA \cite{DBLP:journals/corr/GidarisK15}& \textbf{73.9}&	\textbf{85.5}&	\textbf{82.9}&	\textbf{76.6}&	\textbf{57.8}&	\textbf{62.7}&	\textbf{79.4}&	77.2&	86.6&	\textbf{55.0}&	\textbf{79.1}&	\textbf{62.2}&	87.0&	\textbf{83.4}&	\textbf{84.7}&	78.9&	45.3&	73.4&	65.8&	80.3&	74.0\\
HyperNet\_VGG & 71.4&	84.2&	78.5&	73.6&	55.6&	53.7&	78.7&	\textbf{79.8}&	87.7&	49.6&	74.9&	52.1&	86.0&	81.7&	83.3&	\textbf{81.8}&	\textbf{48.6}&	\textbf{73.5}&	59.4&	79.9&	65.7\\
HyperNet\_SP & 71.3&	84.1&	78.3&	73.3&	55.5&	53.6&	78.6&	79.6&	87.5&	49.5&	74.9&	52.1&	85.6&	81.6&	83.2&	81.6&	48.4&	73.2&	59.3&	79.7&   65.6\\
\rowcolor{Gray}
\textbf{Fast R-CNN + YOLO} & 70.7 & 83.4 & 78.5 & 73.5 & 55.8 & 43.4 & 79.1 & 73.1 & \textbf{89.4} & 49.4 & 75.5 & 57.0 & \textbf{87.5} & 80.9 & 81.0 & 74.7 & 41.8 & 71.5 & 68.5 & \textbf{82.1} & 67.2 \\
MR\_CNN\_S\_CNN \cite{DBLP:journals/corr/GidarisK15}& {70.7}& {85.0}& {79.6}& 71.5& 55.3& {57.7}& 76.0& {73.9}& 84.6& {50.5}& {74.3}& {61.7}& 85.5& 79.9& {81.7}& {76.4}& 41.0& 69.0& 61.2& 77.7& {72.1} \\
Faster R-CNN \cite{ren2015faster}& 70.4&	84.9&	79.8&	74.3&	53.9&	49.8&	77.5&	75.9&	88.5&	45.6&	77.1&	55.3&	86.9&	81.7&	80.9&	79.6&	40.1&	72.6&	60.9&	81.2&	61.5\\
DEEP\_ENS\_COCO &  70.1& 84.0& 79.4& 71.6& 51.9& 51.1& 74.1& 72.1& 88.6& 48.3& 73.4& 57.8& 86.1& 80.0& 80.7& 70.4& {46.6}& 69.6& \textbf{68.8}& 75.9& 71.4 \\
NoC \cite{DBLP:journals/corr/RenHGZ015} &68.8& 82.8& 79.0& 71.6& 52.3& 53.7& 74.1& 69.0& 84.9& 46.9& {74.3}& 53.1& 85.0& {81.3}& 79.5& 72.2& 38.9& {72.4}& 59.5& 76.7& 68.1\\
Fast R-CNN \cite{DBLP:journals/corr/Girshick15}& 68.4 & 82.3 & 78.4 & 70.8 & 52.3 & 38.7 & 77.8 & 71.6 & {89.3} & 44.2 & 73.0 & 55.0 & \textbf{87.5} & 80.5 & 80.8 & 72.0 & 35.1 & 68.3 & 65.7 & 80.4 & 64.2 \\
UMICH\_FGS\_STRUCT& 66.4&	82.9&	76.1&	64.1&	44.6&	49.4&	70.3&	71.2&	84.6&	42.7&	68.6&	55.8&	82.7&	77.1&	79.9&	68.7&	41.4&	69.0&	60.0&	72.0&	66.2\\
NUS\_NIN\_C2000 \cite{dong2014towards}& 63.8 & 80.2 & 73.8 &  61.9 &  43.7 &  43.0 &  70.3 &  67.6 &  80.7 &  41.9 &  69.7 &  51.7 &  78.2 &  75.2 &  76.9 &  65.1 &  38.6 &  68.3 &  58.0 &  68.7 &  63.3 \\
BabyLearning \cite{dong2014towards}&  63.2 &  78.0 &  74.2 &  61.3 &  45.7 &  42.7 &  68.2 &  66.8 &  80.2 &  40.6 &  70.0 &  49.8 &  79.0 &  74.5 &  77.9 &  64.0 &  35.3 &  67.9 &  55.7 &  68.7 &  62.6 \\
NUS\_NIN & 62.4 &  77.9 &  73.1 &  62.6 &  39.5 &  43.3 &  69.1 &  66.4 &  78.9 &  39.1 &  68.1 &  50.0 &  77.2 &  71.3 &  76.1 &  64.7 &  38.4 &  66.9 &  56.2 &  66.9 &  62.7 \\
R-CNN VGG BB \cite{girshick2014rich}&  62.4 &  79.6 &  72.7 &  61.9 &  41.2 &  41.9 &  65.9 &  66.4 &  84.6 &  38.5 &  67.2 &  46.7 &  82.0 &  74.8 &  76.0 &  65.2 &  35.6 &  65.4 &  54.2 &  67.4 &  60.3 \\
R-CNN VGG \cite{girshick2014rich}& 59.2 &  76.8 &  70.9 &  56.6 &  37.5 &  36.9 &  62.9 &  63.6 &  81.1 &  35.7 &  64.3 &  43.9 &  80.4 &  71.6 &  74.0 &  60.0 &  30.8 &  63.4 &  52.0 &  63.5 &  58.7 \\
\rowcolor{Gray}
\textbf{YOLO} &57.9&	77.0&	67.2&	57.7&	38.3&	22.7&	68.3&	55.9&	81.4&	36.2&	60.8&	48.5&	77.2&	72.3&	71.3&	63.5&	28.9&	52.2&	54.8&	73.9&	50.8\\
Feature Edit \cite{shen2014more}&  56.3 &  74.6 &  69.1 &  54.4 &  39.1 &  33.1 &  65.2 &  62.7 &  69.7 &  30.8 &  56.0 &  44.6 &  70.0 &  64.4 &  71.1 &  60.2 &  33.3 &  61.3 &  46.4 &  61.7 &  57.8 \\
R-CNN BB \cite{girshick2014rich}&  53.3 &  71.8 &  65.8 &  52.0 &  34.1 &  32.6 &  59.6 &  60.0 &  69.8 &  27.6 &  52.0 &  41.7 &  69.6 &  61.3 &  68.3 &  57.8 &  29.6 &  57.8 &  40.9 &  59.3 &  54.1 \\
SDS \cite{hariharan2014simultaneous}& 50.7 &  69.7 &  58.4 &  48.5 &  28.3 &  28.8 &  61.3 &  57.5 &  70.8 &  24.1 &  50.7 &  35.9 &  64.9 &  59.1 &  65.8 &  57.1 &  26.0 &  58.8 &  38.6 &  58.9 &  50.7 \\
R-CNN \cite{girshick2014rich}& 49.6 & 68.1 & 63.8 & 46.1 & 29.4 & 27.9 & 56.6 & 57.0 & 65.9 & 26.5 & 48.7 & 39.5 & 66.2 & 57.3 & 65.4 & 53.2 & 26.2 & 54.5 & 38.1 & 50.6 & 51.6 \\
\end{tabularx}
\end{center}
\caption{\small \textbf{\textsc{Pascal} VOC 2012 Leaderboard.} YOLO compared with the full \texttt{comp4} (outside data allowed) public leaderboard as of November 6th, 2015. Mean average precision and per-class average precision are shown for a variety of detection methods. YOLO is the only real-time detector. Fast R-CNN + YOLO is the forth highest scoring method, with a 2.3\% boost over Fast R-CNN.} \vspace{-.3cm}
\label{results}
\end{table*}

YOLO makes far fewer background mistakes than Fast R-CNN. By using YOLO to eliminate background detections from Fast R-CNN we get a significant boost in performance. For every bounding box that R-CNN predicts we check to see if YOLO predicts a similar box. If it does, we give that prediction a boost based on the probability predicted by YOLO and the overlap between the two boxes.

The best Fast R-CNN model achieves a mAP of 71.8\% on the VOC 2007 test set. When combined with YOLO, its mAP increases by 3.2\% to 75.0\%. We also tried combining the top Fast R-CNN model with several other versions of Fast R-CNN. Those ensembles produced small increases in mAP between .3 and .6\%, see Table \ref{combine} for details.

The boost from YOLO is not simply a byproduct of model ensembling since there is little benefit from combining different versions of Fast R-CNN. Rather, it is precisely because YOLO makes different kinds of mistakes at test time that it is so effective at boosting Fast R-CNN's performance.

Unfortunately, this combination doesn't benefit from the speed of YOLO since we run each model seperately and then combine the results. However, since YOLO is so fast it doesn't add any significant computational time compared to Fast R-CNN.

\subsection{VOC 2012 Results}

On the VOC 2012 test set, YOLO scores 57.9\% mAP. This is lower than the current state of the art, closer to the original R-CNN using VGG-16, see Table \ref{results}. Our system struggles with small objects compared to its closest competitors. On categories like \texttt{bottle}, \texttt{sheep}, and \texttt{tv/monitor} YOLO scores 8-10\% lower than R-CNN or Feature Edit. However, on other categories like \texttt{cat} and \texttt{train} YOLO achieves higher performance.

Our combined Fast R-CNN + YOLO model is one of the highest performing detection methods. Fast R-CNN gets a 2.3\% improvement from the combination with YOLO, boosting it 5 spots up on the public leaderboard.

\ifx 1 0
\subsection{Speed}

At test time YOLO processes images at 45 frames per second on an Nvidia Titan X GPU. It is considerably faster than classifier-based methods with similar mAP. Normal R-CNN using AlexNet or the small VGG network take 400-500x longer to process images. The recently proposed Fast R-CNN shares convolutional features between the bounding boxes but still relies on selective search for bounding box proposals which accounts for the bulk of their processing time. YOLO is still around 100x faster than Fast R-CNN. Table \ref{timing} shows a full comparison between multiple R-CNN and Fast R-CNN variants and YOLO.

\begin{table}[h]
\begin{center}
\begin{tabular}{lrrrr}
 & mAP & Test Time & FPS\\
\hline
R-CNN (VGG-16) & 66.0 & 48.2 hr & 0.02 fps \\
FR-CNN (VGG-16) & 66.9 & 3.1 hr & 0.45 fps \\
R-CNN (Small VGG) & 60.2 & 14.4 hr & 0.09 fps \\
FR-CNN (Small VGG) & 59.2 & 2.9 hr & 0.48 fps \\
R-CNN (Caffe) & 58.5 & 12.2 hr & 0.11 fps \\
FR-CNN (Caffe) & 57.1 & 2.8 hr & 0.48 fps \\
YOLO & 63.5 & 110 sec & 45 fps \\
\end{tabular}
\end{center}
\caption{\small \textbf{Prediction Timing.} mAP and timing information for R-CNN, Fast R-CNN, and YOLO on the VOC 2007 test set. Timing information is given both as frames per second and the time each method takes to process the full 4952 image set. The final column shows the relative speed of YOLO compared to that method.}
\label{timing}
\end{table}
\fi

\subsection{Generalizability: Person Detection in Artwork}

\begin{figure*}
\centering
\begin{subfigure}[b]{.45\textwidth}
    \centering
    \includegraphics[width=\textwidth]{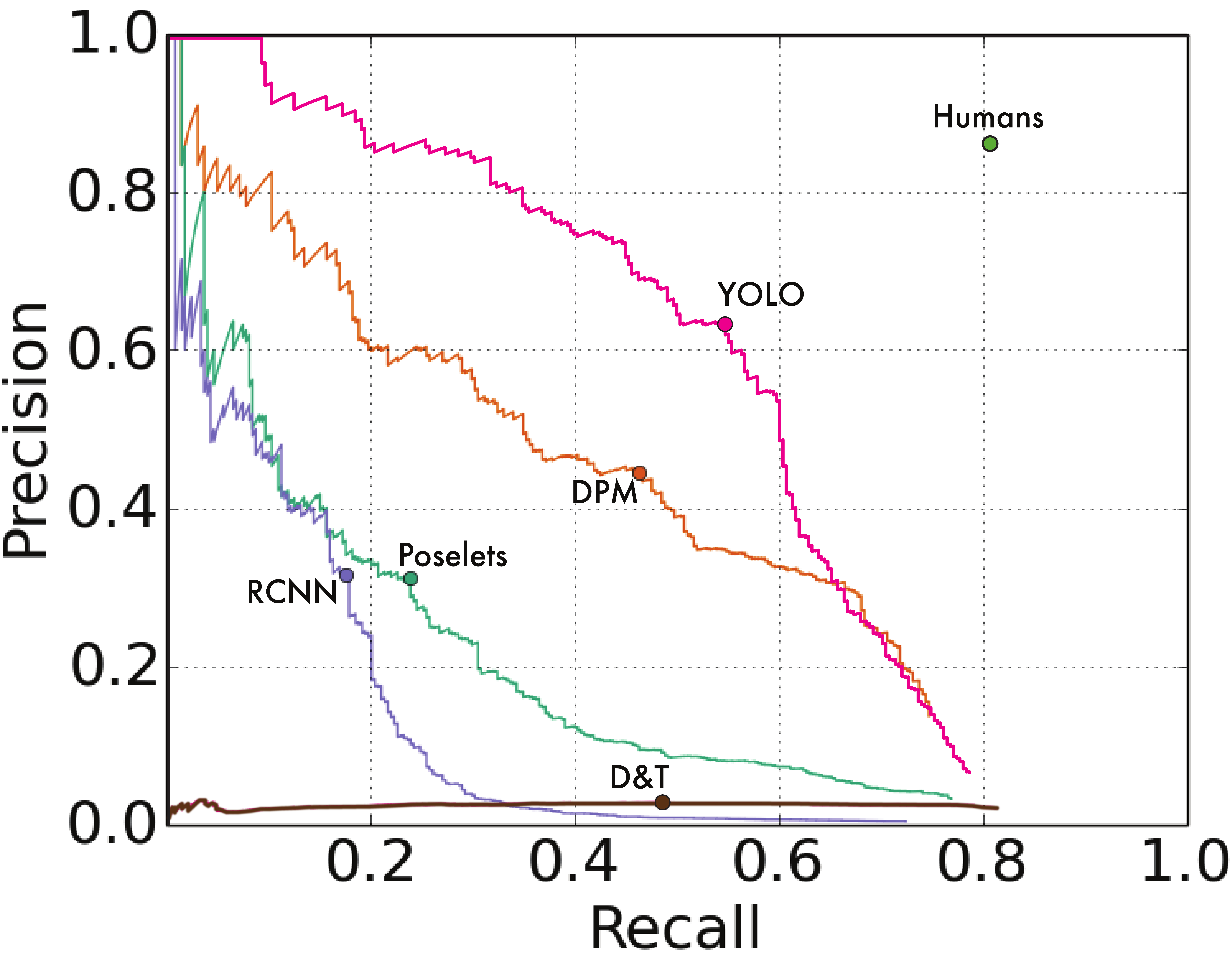}
    \caption{\small Picasso Dataset precision-recall curves.}
\end{subfigure}%
\begin{subfigure}[b]{.55\textwidth}
    \centering
    \begin{tabular}{l|r|rr|r}
& VOC 2007 & \multicolumn{2}{c|}{Picasso} & People-Art\\
 & AP & AP & Best $F_1$ & AP\\
\hline
\textbf{YOLO} & \textbf{59.2} & \textbf{53.3} & \textbf{0.590} & \textbf{45}\\
R-CNN & 54.2 & 10.4 & 0.226 & 26\\
DPM & 43.2 & 37.8 & 0.458 & 32\\
Poselets \cite{BourdevMalikICCV09} & 36.5 & 17.8 & 0.271 \\
D\&T \cite{dalal2005histograms} & - & 1.9 & 0.051 \\
\end{tabular}
\caption{\small Quantitative results on the VOC 2007, Picasso, and People-Art Datasets. The Picasso Dataset evaluates on both AP and best $F_1$ score.}
\end{subfigure}
\caption{\small \textbf{Generalization results on Picasso and People-Art datasets.}}
\label{art}
\end{figure*}

\begin{figure*}[t]
\begin{center}
    \includegraphics[width=\linewidth]{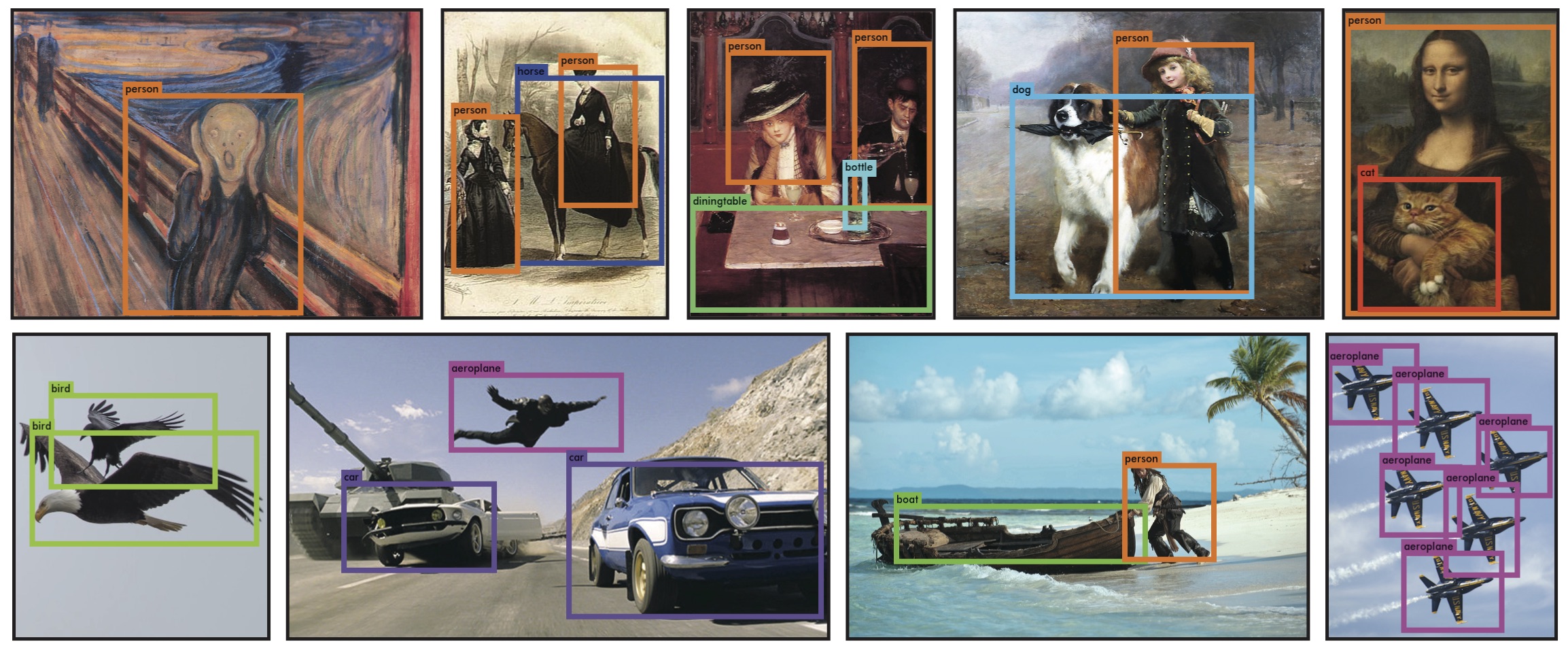}
\end{center}
   \caption{\small \textbf{Qualitative Results.} YOLO running on sample artwork and natural images from the internet. It is mostly accurate although it does think one person is an airplane.}
\label{images}
\end{figure*}

Academic datasets for object detection draw the training and testing data from the same distribution. In real-world applications it is hard to predict all possible use cases and the test data can diverge from what the system has seen before \cite{cai2015cross}. 
%We want our detector to learn robust representations so that they can generalize to new domains.
We compare YOLO to other detection systems on the Picasso Dataset \cite{ginosar2014detecting} and the People-Art Dataset \cite{cai2015cross}, two datasets for testing person detection on artwork.
%Models are trained on subsets of the VOC data and then run on artwork to detect people.

Figure \ref{art} shows comparative performance between YOLO and other detection methods. For reference, we give VOC 2007 detection AP on \texttt{person} where all models are trained only on VOC 2007 data. On Picasso models are trained on VOC 2012 while on People-Art they are trained on VOC 2010.

R-CNN has high AP on VOC 2007. However, R-CNN drops off considerably when applied to artwork. R-CNN uses Selective Search for bounding box proposals which is tuned for natural images. The classifier step in R-CNN only sees small regions and needs good proposals.

DPM maintains its AP well when applied to artwork. Prior work theorizes that DPM performs well because it has strong spatial models of the shape and layout of objects. Though DPM doesn't degrade as much as R-CNN, it starts from a lower AP.

YOLO has good performance on VOC 2007 and its AP degrades less than other methods when applied to artwork. Like DPM, YOLO models the size and shape
of objects, as well as relationships between objects and where objects commonly appear. Artwork and natural images are very different on a pixel level but they are similar in terms of the size and shape of objects, thus YOLO can still predict good bounding boxes and detections.

\section{Real-Time Detection In The Wild}

YOLO is a fast, accurate object detector, making it ideal for computer vision applications. We connect YOLO to a webcam and verify that it maintains real-time performance, including the time to fetch images from the camera and display the detections.

The resulting system is interactive and engaging. While YOLO processes images individually, when attached to a webcam it functions like a tracking system, detecting objects as they move around and change in appearance. A demo of the system and the source code can be found on our project website: \url{http://pjreddie.com/yolo/}.

\section{Conclusion}

We introduce YOLO, a unified model for object detection. Our model is simple to construct and can be trained directly on full images. Unlike classifier-based approaches, YOLO is trained on a loss function that directly corresponds to detection performance and the entire model is trained jointly.

Fast YOLO is the fastest general-purpose object detector in the literature and YOLO pushes the state-of-the-art in real-time object detection. YOLO also generalizes well to new domains making it ideal for applications that rely on fast, robust object detection.

\noindent\textbf{Acknowledgements:} This work is partially supported by ONR N00014-13-1-0720, NSF IIS-1338054, and The Allen Distinguished Investigator Award.

%A cell is only responsible for objects centered in that cell so we restrict the coordinates to fall within that cell. This imposes a strong spatial constraint on the predictions which has both benefits and drawbacks. Cells rarely predict multiple bounding boxes for the same object.

%Conversely, this spatial constraint limits the number of nearby objects that our model can predict. If two objects fall into the same cell our model can only predict one of them. Our model struggles with small objects that appear in groups, such as flocks of birds.

%Finally, our loss function treats errors the same in small bounding boxes versus large bounding boxes. A small error in a large box is generally benign but a small error in a small box has a much greater effect on IOU. Our main source of error is incorrect localizations.

\pagebreak
{\small
\bibliographystyle{ieee}
\bibliography{egbib}
}

\end{document}